\documentclass[11pt,a4paper]{article}
\usepackage[hyperref]{acl2021}
\usepackage{times}
\usepackage{latexsym}

\usepackage{subfigure}
\usepackage{makecell}
\usepackage{arydshln}
\usepackage{enumitem}
\usepackage{graphicx}

\usepackage{multirow}

\usepackage{microtype}

\aclfinalcopy 

\newcommand{\dataname}{\textsc{DocNLI}}

\title{\dataname: A Large-scale Dataset for Document-level \\Natural Language Inference}

\author{Wenpeng Yin$^1$, 
\textbf{Dragomir Radev$^{1,2}$}  and \textbf{Caiming Xiong$^1$} \\
  $^1$Salesforce Research and $^2$Yale University\\
  {\tt
    wyin@salesforce.com}}
    
\date{}

\begin{document}
\maketitle
\begin{abstract}
Natural language inference (NLI) is formulated as a unified framework for solving various NLP problems such as relation extraction, question answering, summarization, etc. It has been studied intensively in the past few years thanks to the availability of large-scale labeled datasets. However, most existing  studies focus on merely sentence-level inference, which limits the scope of NLI's application in downstream NLP problems. This work presents \dataname\enspace--- a newly-constructed large-scale dataset for document-level NLI. \dataname\enspace is transformed from a broad range of NLP problems and covers multiple genres of text. The premises always stay in the document  granularity, whereas the hypotheses vary in length from single sentences to passages with hundreds of words. Additionally, \dataname\enspace has pretty limited artifacts\footnote{NLI ``artifacts" are some label-specific biases (often) in the hypotheses; they can  indicate which NLI class a hypothesis  belongs to even without looking at the premise.} which unfortunately widely exist in some popular sentence-level NLI datasets. Our experiments demonstrate that, even without fine-tuning,  a model pretrained on \dataname\enspace shows promising performance on popular sentence-level benchmarks, and generalizes well to out-of-domain NLP tasks that rely on inference at document granularity. Task-specific fine-tuning can  bring further improvements. Data, code and pretrained models can be found at \url{https://github.com/salesforce/DocNLI}.
\end{abstract}

\section{Introduction}\label{sec:intro}
A fundamental challenge of natural language processing (NLP) lies in the variability of semantic expression, where the same meaning can be conveyed by, or inferred from, different pieces of text \cite{DBLPDaganDMR10}. This phenomenon  gives rise to the many-to-many mapping between textual expressions and meanings. Many NLP problems, such as information extraction, question answering, document summarization and machine translation, desire a system for this variability phenomenon so as to figure out that a particular  meaning can be inferred from distinct text strings \cite{DBLPDaganDMR10}. Natural language inference (a.k.a textual entailment \cite{DBLPDaganGM05}) acts as a unified framework to study those NLP problems by casting the background text as a premise and the text of target meaning as a hypothesis. Then, a good NLI recognizer  can be considerably translated to a well-performing system regarding respective NLP tasks.

NLI was first studied in \cite{DBLPDaganGM05}. Research  in the early stages was mostly driven by the PASCAL Recognizing Textual Entailment (RTE) challenges which are annual competitions with benchmark datasets released.  In the past few years, the study of NLI has moved forward rapidly along with  the construction of large-scale datasets, such as SNLI \citep{DBLPBowmanAPM15}, the science domain SciTail \citep{DBLPKhotSC18} and multi-genre MNLI \citep{DBLPWilliamsNB18}, etc.

However, some NLI datasets may not be suitable any more for solving downstream NLP problems since they were commonly crowdsourced in isolation from any end task \footnote{Except for  RTE and SciTail} \cite{DBLPKhotSC18}. In addition, most NLI datasets and studies paid attention merely to sentence-level inference --- both the premises and hypotheses are single (and usually short)  sentences. This makes them unsuitable for other open-ended NLP problems. For example, to verify the factual correctness of a document summary, sentence-level NLI systems cannot be of much help \cite{DBLP12840}. Considering the fact-checking task FEVER \cite{DBLPhorneVCM18} as another example, in order to figure out the truth value of a claim against a Wikipedia article, NLI has to be done on individual sentences instead of using the whole article as the premise. In short, some NLP tasks require the  reasoning of NLI to go beyond the sentence granularity, regarding both the premise and the hypothesis.

In this work, we introduce \dataname, a large-scale dataset for document-level NLI. It is constructed  by reformatting some mainstream NLP tasks, including question answering and document summarization, and integrating existing NLI in which  the premises may be longer than single sentences. \dataname\enspace has the following characteristics:

\begin{itemize}[leftmargin=3.5mm]
\setlength\itemsep{0.5mm}
    \item \dataname\enspace is highly related with end NLP tasks. 
    A well-performing system to \dataname\enspace  is expected to  throw light on addressing other NLP challenges. 
    \item Premises always have more than one sentence; the majority are natural documents such as  news articles. Hypotheses cover a variety of lengths, ranging from a single sentence to  a  document with hundreds of words. By this setting, we hope the systems can learn to deal with future applications that need to infer the truth value of a piece of text regardless of its length.
    \item In contrast to some existing sentence-level NLI datasets, \dataname\enspace has pretty limited artifacts. We present a novel approach to disconnect the potential artifacts with the NLI task itself; a ``hypothesis-only'' baseline has difficulties in  discovering some spurious correlations. 
\end{itemize}


In experiments, we will show that a RoBERTa \cite{DBLP11692} system pre-trained on \dataname\enspace demonstrates promising performance on conventional sentence-level NLI benchmarks such as MNLI and SciTail, and generalizes well to out-of-domain NLP tasks (e.g., fact-checking and multi-choice question answering) that necessitate document-level inference. Task-specific fine-tuning can further improve the performance and achieve new state of the art for some end tasks.


\begin{table*}[t]
\setlength{\tabcolsep}{2pt}
  \centering
  \begin{tabular}{l||c|c|c|c}
&  original task  & domain  & premise length & hypothesis length\\\hline\hline
\multirow{2}{*}{ANLI} & \multirow{2}{*}{NLI} & various  & multi-sentence  & single sentence\\
 & & (wiki, news, etc.) &  (20$\sim$94 words) & (4$\sim$18 words)\\\hline
 


 \multirow{2}{*}{SQuAD} & \multirow{2}{*}{QA} & \multirow{2}{*}{wiki}  & paragraph  & single sentence\\
 & &  & (27$\sim$237 words) & (6$\sim$22 words)\\\hline

  DUC & \multirow{2}{*}{summarization} & \multirow{2}{*}{news}  & doc.  & multi-sent \\
 (2001)& &  & (124$\sim$879 words) & ($80\sim$100 words)\\\hline
 
 CNN/Daily & \multirow{2}{*}{summarization} & \multirow{2}{*}{news}  & doc.  & 3$\sim$4 sent. \\
 Mail& &  & (247$\sim$652 words) & (40$\sim$50 words)\\\hline

  \multirow{2}{*}{Curation} & \multirow{2}{*}{summarization} & \multirow{2}{*}{news} & doc.  & multi-sent\\
 & &  &  (229$\sim$842 words) & (64$\sim$279 words)\\\hline\hline
\end{tabular}
\caption{Data resources that are reformatted into \dataname. }\label{tab:dataOrigin}
\end{table*}

\section{Related Work}
To our knowledge, document-level NLI has attracted very little ink in the community, possibly  because of the lack of labeled datasets. In this section, we mainly describe some prior NLI datasets that share some spirits with our \dataname.  

\paragraph{End-task driven.} As  mentioned in Section \ref{sec:intro}, the RTE series were driven by downstream NLP tasks such as information retrieval, information extraction, question answering,  and summarization. MCTest \cite{DBLPichardsonBR13} is a question answering task in which a paragraph is given as background knowledge, then each question is paired with a positive answer and some negative answers. The MCTest benchmark released an NLI version of this corpus by treating the whole paragraph as a premise and combining the question and answer candidates as hypotheses.   SciTail \cite{DBLPKhotSC18} is also derived from the end QA task of answering multiple-choice school-level science questions. Unlike MCTest, the premises in SciTail are single sentences selected by an information retrieval approach. By casting an end NLP task as NLI, a good NLI recognizer therefore can be directly turned into a well-performing system for that NLP task. This can be even more attractive if we can learn a generalizable NLI system to solve some NLP problems that have limited annotations.

\paragraph{Going beyond  the sentence granularity.} The premises in MCTest are  paragraphs, but MCTest has pretty limited size. 
\newcite{DBLP02922} tried to convert the question answering benchmark SQuAD \cite{DBLPRajpurkarZLL16} into an NLI format by treating the paragraph as a premise and using a neural network to generate a hypothesis sentence given the question and answer span as inputs.   \newcite{DBLP12840} created a (document, sentence) pair data ``FactCC'' to train a classifier for checking the factual correctness of single sentences in automatically generated summaries. FactCC is specific to the target summarization benchmark dataset, so it is unclear how well FactCC can generalize to other summarization benchmarks and other NLP problems. In addition, only single sentences act as hypotheses. Nevertheless, that literature exactly showed that document-level NLI, especially the inference of document-level hypotheses, is highly desirable. ANLI \cite{DBLPabs191014599} also  gather multi-sentence as premises. However, the sentence sizes in ANLI premises are pretty  limited and the hypotheses in ANLI are single sentences consistently.

To our knowledge, our \dataname\enspace is the first dataset that uses hypotheses longer than single sentences, and stays closely with end NLP tasks.

\section{Data Creation}
\textit{What kind of document-level NLI dataset is preferred?} (i) We want  the premise is a paragraph or even a document, and the hypotheses cover a large range of granularity: from a single sentence to a longer paragraph (e.g., 250 words); (ii) Diverse domains; (iii) No severe artifacts; for example, we do not include the hypotheses that can be easily found ``grammatically incorrect'' by well-trained language models such as BERT \cite{DBLP04805}.
\subsection{Data Preprocessing}\label{sec:datapreprocessing}

Table \ref{tab:dataOrigin} lists all the resources that we use to create  \dataname. Briefly,  \dataname\enspace combines and reformats five existing NLP benchmarks: 
adversarial NLI (ANLI) \cite{DBLPabs191014599}, the question answering benchmark SQuAD \cite{DBLPRajpurkarZLL16} and three summarization benchmarks (DUC2001\footnote{\url{https://www-nlpir.nist.gov/projects/duc/guidelines/2001.html}}, CNN/DailyMail \cite{DBLPNallapatiZSGX16}, and Curation\footnote{\url{https://github.com/CurationCorp/curation-corpus}} \cite{curationcorpusbase2020}). Next, we describe how each data resource is integrated into \dataname.

\paragraph{ANLI to \dataname.} ANLI is a large-scale NLI dataset collected via an iterative, adversarial human-and-model-in-the-loop procedure. In each round, the best-performing model from the previous round is selected and then human annotators are asked to write ``hard'' examples that this model misclassifies. They always choose \textit{multi-sentence paragraphs as premises and write single sentences as hypotheses}. Then a part of those ``hard'' examples join the training set so as to learn a stronger model for the next round. The remaining  ``hard'' examples act as dev/test sets correspondingly. Totally three rounds were accomplished for ANLI construction. In the end, ANLI has train/dev/test  sizes as 162,865/3200/3200 with three classes ``entail'', ``neutral'' and ``contradict''. 

We keep premise-hypothesis pairs in ANLI unchanged, but unify the two classes ``neutral'' and ``contradict'' into a new class ``not\_entail''.

\paragraph{SQuAD to \dataname.} SQuAD  is a QA dataset in which a multi-sentence paragraph is accompanied by a couple of questions; each question has a text span from the paragraph as its answer. \newcite{DBLP02922} converted SQuAD into NLI format by reformatting the question-answer pairs into  declarative sentences (QA2D) by neural networks. The resulting sentences containing correct (resp. incorrect) answers are entailed (resp. not\_entail) by the paragraph. Human evaluation was conducted to make sure those declarative sentences have high quality on three criteria: grammaticality, naturalness, and completeness. In addition, \newcite{DBLP02922} replicated some statistical analyses showing that this QA2D dataset does not have clear artifacts as SNLI or MNLI.  In this work, we directly use this QA2D dataset and re-split it into train/dev/test by 50k/7,236/8,275.

\begin{table*}[t]
\setlength{\tabcolsep}{1pt}
  \centering
  \small
  \begin{tabular}{ll|l}
  \hline\hline
\multicolumn{2}{c|}{\multirow{11}{*}{doc}} & Petrofac shares surged on Wednesday following reports that the Serious Fraud Office has abandoned a criminal \\
&& investigation  into three businessmen who were accused of paying brides in the energy industry. The SFO had\\
&&    been probing claims that Unaoil - a Monaco-based consultancy that worked with Petrofac, primarily in Kazakhstan\\
&&   between 2002 and 2009 - had paid multimillion pound brides to land contracts in the oil and gas industry.\\
&&  But The Guardian cited sources earlier as saying that the SFO has dropped the investigation  into the trio.\\
&&  Compliance industry newsletter MLex was the first to report the news, saying on Tuesday that the probe had been\\
&&    halted after three years. The SFO launched an investigation into Petrofac in May 2017 as part of a wider probe\\
&&   into Unaoil. In February 2019, David Lufkin, Petrofac's former global head of sales, pleaded guilty to 11 counts of\\
&& bribery linked to contracts worth more than \$730m in Iraq and \$3.5bn in Saudi Arabia. SFO spokesman Adam \\
&& Lilley  said the Unaoil investigation "remains active and is ongoing". ``We do not comment on ongoing\\
&&  investigations," he said. [$\cdots$]\\\hline

   \multicolumn{2}{c|}{\multirow{3}{*}{\makecell{real\\ summ.} }} &  The Serious Fraud Office has reportedly dropped a criminal investigation into three businessmen who had been\\
&&  accused of conspiring to make corrupt payments to secure contracts in Iraq. The SFO launched an investigation\\
&&  into Petrofac in May 2017 as part of a wider probe into Monaco-based oil consultancy Unaoil.\\\hline

   \multirow{9}{*}{\rotatebox{90}{fake summaries}}&\multirow{3}{*}{\makecell{word\\ repl.} }& The Serious \textcolor{red}{financial} Office has reportedly \textcolor{red}{launched} a criminal investigation into three businessmen who had\\
  &&  been accused of conspiring to make corrupt payments to \textcolor{red}{oil} contracts in Iraq. The SFO launched an investigation\\
  &&  into \textcolor{red}{corruption} in May 2017 as part of a wider \textcolor{red}{investigation} into Monaco-based \textcolor{red}{financial} consultancy \textcolor{red}{firms}.\\\cdashline{2-3}
  
  &\multirow{3}{*}{\makecell{entity\\ repl.} }& \textcolor{red}{Unaoil} has reportedly dropped a criminal investigation into three businessmen who had been accused of conspiring\\
  &&   to make corrupt payments to secure contracts in \textcolor{red}{Monaco} . The SFO launched an investigation into \textcolor{red}{Monaco} in\\
  &&    May 2017 as part of a wider probe into \textcolor{red}{Petrofac}-based  oil consultancy \textcolor{red}{The Serious Fraud Office}.\\\cdashline{2-3}

  &\multirow{3}{*}{\makecell{sent\\ repl.} }& The Serious Fraud Office has reportedly dropped a criminal investigation into three businessmen who had been\\
  &&   accused of conspiring to make corrupt payments to secure contracts in Iraq. \textcolor{red}{A spokesman for the SFO said it was}\\
  &&    \textcolor{red}{  ``unable to confirm} \textcolor{red}{ or deny'' that an inquiry had taken place.}\\\hline

\end{tabular}
\caption{An example of the Curation summarization dataset shows  the original document, and the real summary written by humans. We used ``word replacement'', ``entity replacement'' and ``sentence replacement'' to form three types of ``fake'' summaries against the document. Texts in red are substitutes.}\label{tab:summarychanceexample}
\end{table*}

\paragraph{Summarization to \dataname.} Here we introduce the basics of the three summarization datasets (DUC2001, CNN/DailyMail and Curation), and  explain how we convert them into \dataname\enspace in a unified approach. 

\textbullet\enspace The DUC series are some of the earliest benchmarks for studying automatic document summarization. 
DUC2001 is on generic,   single-document   summarization in the news domain. There are totally 600 documents along with human-written reference summaries of approximately 100 words.  We split those document-summary pairs into train/dev/test by size of 400/50/150.

\textbullet\enspace CNN/DailyMail was gathered from news articles in \textit{CNN} and \textit{Daily Mail} websites; each article is paired with 3 to 4 sentences of abstractive summary bullets generated by humans. CNN/DailyMail has 286,817/13,368/11,487 article-summary pairs in train/dev/test. The source articles in the training set have 766 words spanning 29.74 sentences on  average while the summaries consist of averagely 3.72 sentences.

\textbullet\enspace Curation is a recent summarization dataset with  40,000 professionally-written summaries of news articles. We split it into train/dev/test as 20K/7K/13K.

All  three summarization datasets align the documents with the human-written reference summaries. This enables us to obtain ``entail'' pairs of (document, reference summary). The remaining challenge lies in how to generate ``not\_entail'' pairs.

We adopt  three types of manipulations on the ``reference'' (also referred as ``real'') summaries.

\textbullet\enspace \textit{Word replacement}. We mask eight words whose part-of-speech tags are among \{``VERB'', ``NOUN'', ``PROPN'', ``NUM''\} by spaCy toolkit\footnote{\url{https://spacy.io}}, then use BERT to predict them. The most likely predicted word is used to replace a masked one. After word replacements, the resulting text is our ``fake'' summary.

\textbullet\enspace \textit{Entity replacement}. We use spaCy for named entity recognition (NER). For an entity which is the only one of a specific NER type in the real summary, we search for a different entity with the same type from the document to replace it; otherwise, it will be replaced by the entity of the same type in the real summary. We do this operation for five entities. We skip entity-level manipulation for  the instances that have fewer than five  detected entities.  After entity replacement, we get a ``fake'' summary.

\textbullet\enspace \textit{Sentence replacement}. From the real summary, we randomly select a sentence, then forward its left context to CTRL \cite{DBLP05858}, a state-of-the-art controllable text generator, to generate a new sentence which is used to replace the selected sentence. This operation generates a new ``fake'' summary.

Table \ref{tab:summarychanceexample} illustrates a (document, real summary) pair in the Curation dataset, and the three types of ``fake'' summaries we generated. 

\begin{table}[t]
  \centering
  \begin{tabular}{lc|cc}
 && entail& not\_entail\\\hline\hline
 
\multirow{4}{*}{\rotatebox{90}{\dataname}} & \multirow{3}{*}{\makecell{raw\\ \dataname} }   & ANLI &  ANLI\\
 && SQuAD & SQuAD\\
 && \{($D$, $R$)\} & \{($D$, $F_i$)\} \\\cdashline{2-4}
 &added pairs  & \{($F_i^+$, $F_i$)\} & \{($F_i$, $R$)\}\\\hline
\end{tabular}
\caption{$D$: a document in summarization benchmarks; $R$: a real summary; $F_i$: a fake summary derived from $R$ ($i$=1$\cdots$ $n$); $F_i^+$: Using CTRL to insert a generated sentence between a random pair of consecutive sentences in the $F_i$, in a way similar to what we described as ``sentence replacement'' in Section \ref{sec:datapreprocessing}.  \dataname's training set  is the combination of raw-\dataname\enspace and those added pairs; \dataname's dev and test sets \emph{do not have trivial pairs \{($F_i^+$, $F_i$)\}}. }\label{tab:addpairs}
\end{table}

\subsection{Mitigating Artifacts in \dataname}
In  Section \ref{sec:datapreprocessing}, we transformed these NLI, QA and summarization datasets  to satisfy the format of \dataname. We refer this resulting dataset as \textit{raw-\dataname}. In consideration of the common artifacts in some popular sentence-level NLI benchmarks \cite{DBLPGururanganSLSBS18,DBLPPoliakNHRD18,DBLPTsuchiya18}, we  tried a ``hypothesis-only'' baseline based on RoBERTa on this \textit{raw-\dataname}. Surprisingly, this baseline indeed obtains non-trivial performance.  This means that RoBERTa can still learn some label-specific biases from the hypotheses, even though we have tried hard to make the ``fake'' summaries coherent and natural.

Nevertheless, this does not mean we have failed to build a robust \dataname\enspace dataset. The surprising behavior of ``hypothesis-only'' in \textit{raw-\dataname} indicates that the BERT classifier can easily recognize the summary is ``real vs. fake'', but ``real vs. fake'' is not the same concept as ``entail vs. not\_entail'' defined in the NLI framework. This is because a ``fake'' one can still be ``entail''-ed if the premise has proper information; and a ``real'' one can also be ``not\_entail'' if the premise does not contain necessary clues for inferring it.

For convenience, we use $D$ as a document, $R$ as the real summary, and \{$F_1$, $F_2$, $\cdots$, $F_n$\} as the $n$ fake summaries derived from $R$. To ensure the model can learn exactly what ``entail vs. not\_entail'' is rather than be misled by the manipulations that yield those ``fake'' text pieces, as Table \ref{tab:addpairs} demonstrates, we prepare the following pairs to extend the \textit{raw-\dataname} and get our final \dataname:
\begin{itemize}[leftmargin=3.5mm]
\setlength\itemsep{0.5mm}
    \item Adding pairs ($F_i^+$, $F_i$), $i=1, \cdots, n$, for  class ``entail''. Here $F_i^+$ has one more sentence  than  $F_i$, inserted by CTRL,  as described in ``sentence replacement'' in Section \ref{sec:datapreprocessing} (here we do insertion rather than replacement). The goal  is to let the system know that a fake summary can also be a positive hypothesis in NLI, if its premise covers  necessary information.
    \item Adding a single pair ($F_i$, $R$) for  class ``not\_entail''. This means the original real summary can also be a negative hypothesis if it includes mis-matching information with its premise. $F_i$ is randomly chosen from the set \{$F_1$, $\cdots$, $F_n$\}.
\end{itemize}

By adding above two sorts of pairs, we want to disconnect the concept of ``real vs. fake'' from ``entail vs. not\_entail'', letting the system learn the essence of NLI. Both the ``real'' and ``fake'' summaries  have the same number of instances of being ``entail'' and ``not\_entail'' in the extended dataset. 

\emph{It is worth mentioning that since the instances ``($F_i^+$, $F_i$) $\rightarrow$ entail'' are very trivial to be recognized, we  add them in the training set only.}

\begin{table}[t]
  \centering
  \begin{tabular}{l|ccc}
 & train&  dev  & test\\\hline\hline
entail. &  466,653 & 28,890 &  33,128\\
not\_entail &475,661 & 205,368 &  233,958\\\hline
sum &942,314 & 234,258 & 267,086 \\\hline

\end{tabular}
\caption{Data sizes of \dataname.}\label{tab:data4task1}
\end{table}


Table \ref{tab:data4task1} lists the sizes of \dataname\enspace for train/dev/test in each class. The training set is roughly balanced, while approximately 12\% examples in dev and test belong to ``entail''. F1 is the  evaluation metric.

\begin{figure}[t]
\centering
\includegraphics[width=7cm]{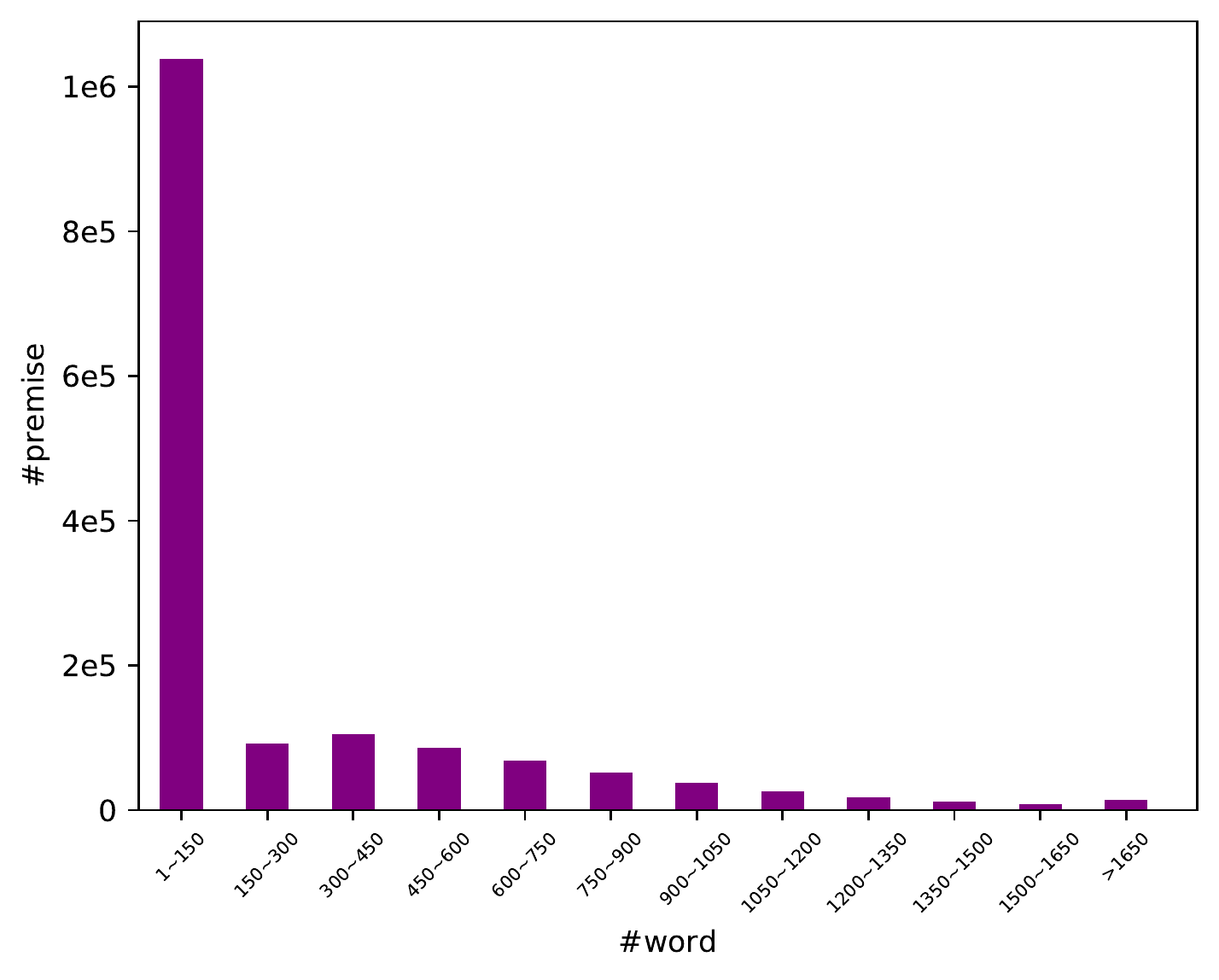}
\caption{\#premise vs. \#words in \dataname
} \label{fig:docnlipremise}
\end{figure}
\begin{figure}[t]
\centering
\includegraphics[width=7cm]{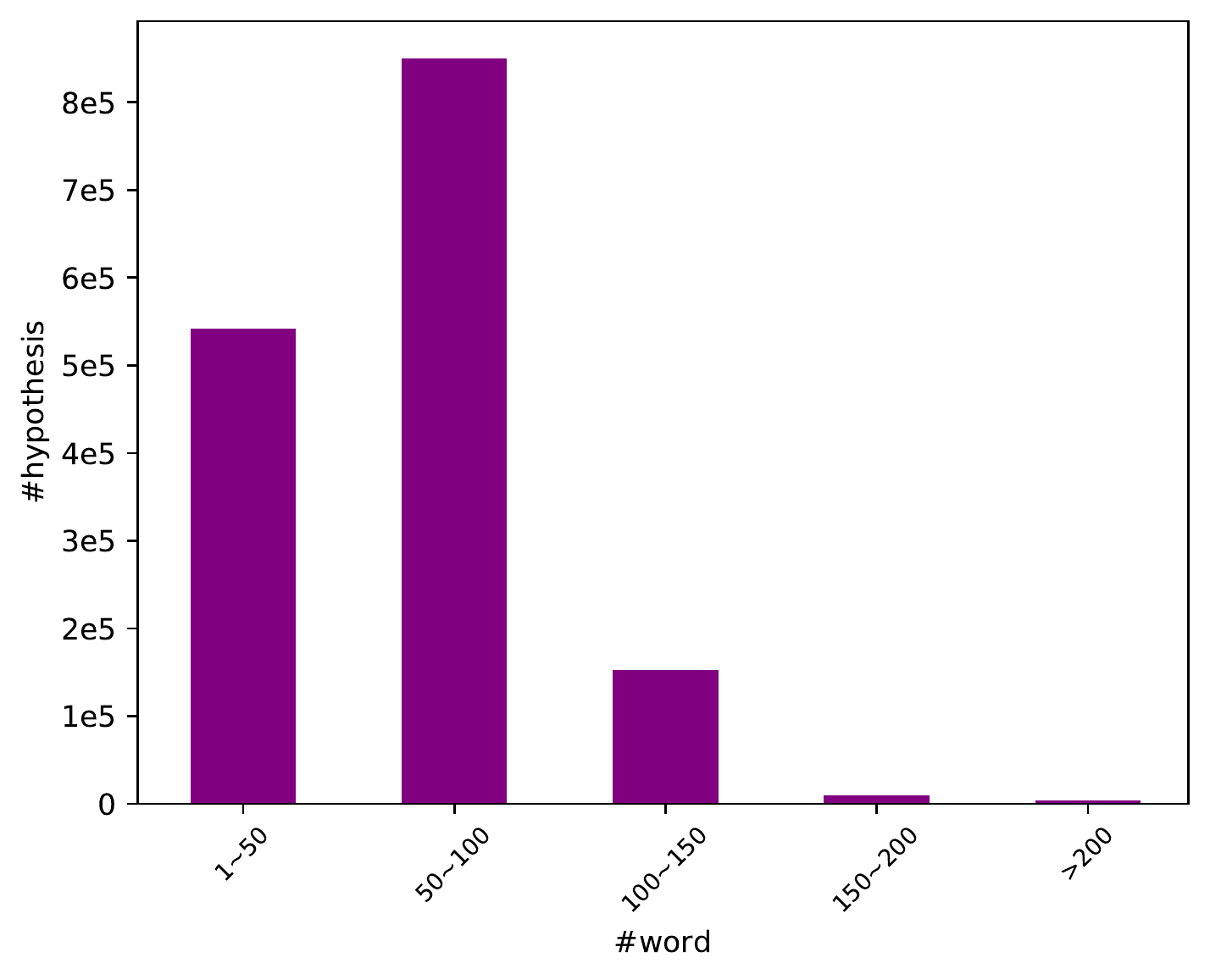}
\caption{\#hypothesis vs. \#words in \dataname. The hypotheses in our new data \dataname\enspace are mostly longer than single sentences; this is one key difference with some related datasets.
} \label{fig:docnlihypo}
\end{figure}

Figures \ref{fig:docnlipremise}-\ref{fig:docnlihypo} illustrate the length distributions of premises and hypotheses in \dataname. Because the majority of hypotheses have fewer than 150 words, and real/fake summaries also act as premises in \dataname, as reported in Table \ref{tab:addpairs}, therefore, the majority of premises stay within the length limit of 150 words, shown in Figure \ref{fig:docnlipremise}. Still, there are a large amount of premises whose lengths are within the range of [150, 900] words.



\subsection{Human Verification}
\dataname\enspace covers examples derived from ANLI, SQuAD and three summarization datasets. Here, we only conduct human verification for the pairs derived from summarization, especially for those ``fake'' summaries, to get some clues to answer two questions: (i) Are those ``fake'' summaries indeed incorrect given the original document? (ii) Do those ``fake'' summaries look natural? By ``natural'' we mean the text should have no major grammar errors, and no unrelated text spans that make the whole text piece look over uncoordinated. 

The authors of this work manually checked 200 random ``fake'' examples, among which none is  true given the same document as the ``real'' summary. This is mainly because we replaced relatively a lot from the original real summaries. 

However, some minor grammar issues inevitably exist. Take the following text piece as an example:

``\textit{\underline{WeWork Companies LLC} (replace: ``WeWork'') has announced plans to hold a conference call on \underline{\textcolor{red}{2025}} for holders of its 7.875\% Senior Notes due \underline{\textcolor{red}{26 August}} to discuss its \underline{Notes} (replace: ``Q2'') results. Securities analysts and market-making financial institutions can also register for access. The call is scheduled for \underline{12:00 P.M.} (replace: ``noon Eastern Time'').}''

This example has five entities that are substitutes, all underlined. If a substitute comes from the premise document, we use ``(replace: XX)'' to denote the entity that was there. The two entities (NER type ``date''), in red, replaced each other: ``2025'' and ``26 August'', which makes the new text ``[$\cdots$] on 2025 [$\cdots$]'' grammatically incorrect.

\begin{figure}[t]
\centering
\includegraphics[width=7cm]{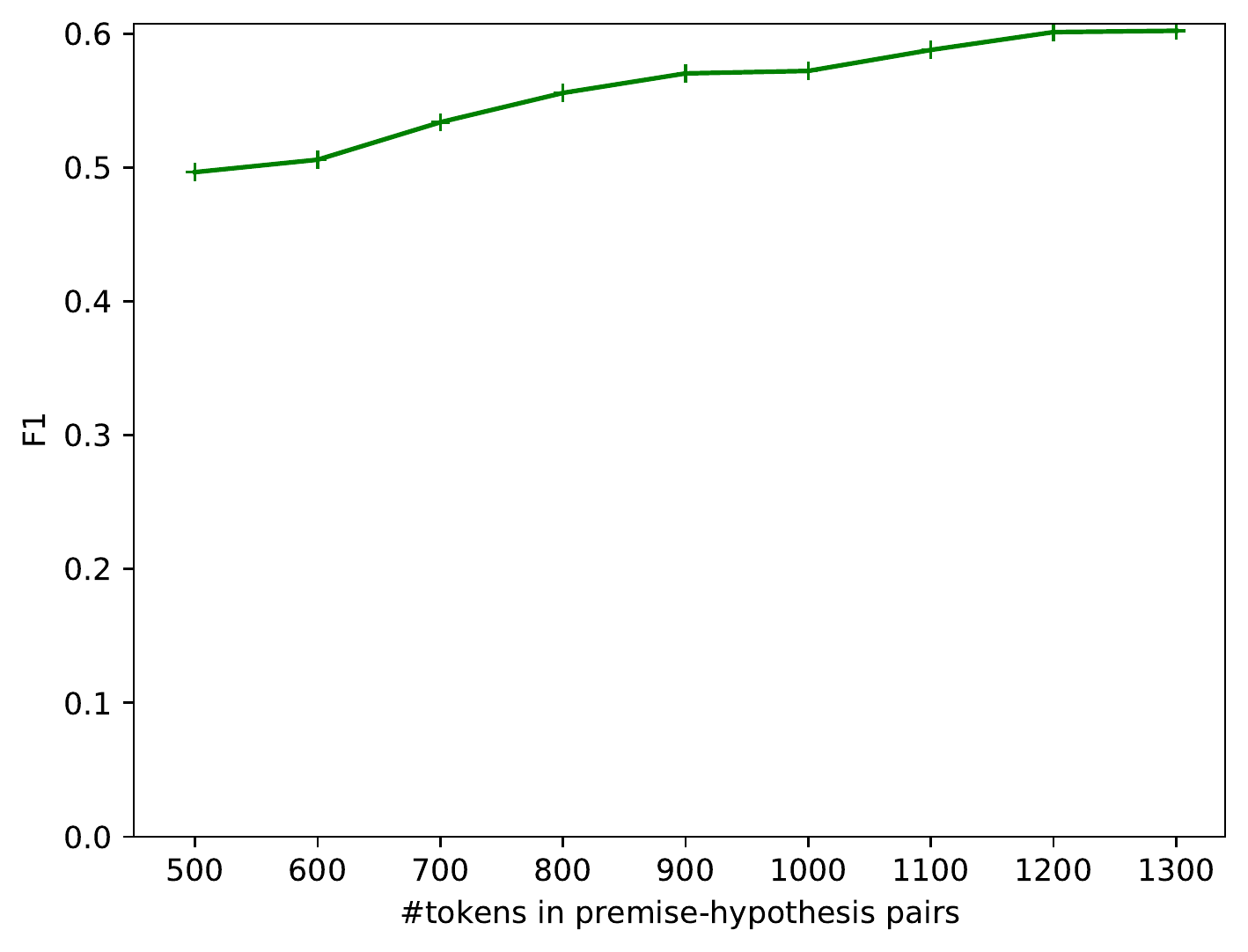}
\caption{ Longformer F1 vs. \#tokens in \dataname\enspace dev set.
} \label{fig:token2f1}
\end{figure}


\begin{table}[t]
  \centering
  \begin{tabular}{l|cc}
&  dev  & test\\\hline\hline
Random & 19.75& 19.91\\
Hypothesis-only & 21.89 & 22.02 \\
Longformer-base & 46.18 &44.42\\
Roberta-large & 63.05 & 61.20\\\hline\hline
\end{tabular}
\caption{F1 scores on \dataname. }\label{tab:results4task1}
\end{table}



\begin{figure*}
\centering
\subfigure[F1 vs. pair length] { \label{fig:Docf1pair}
\includegraphics[width=7.5cm]{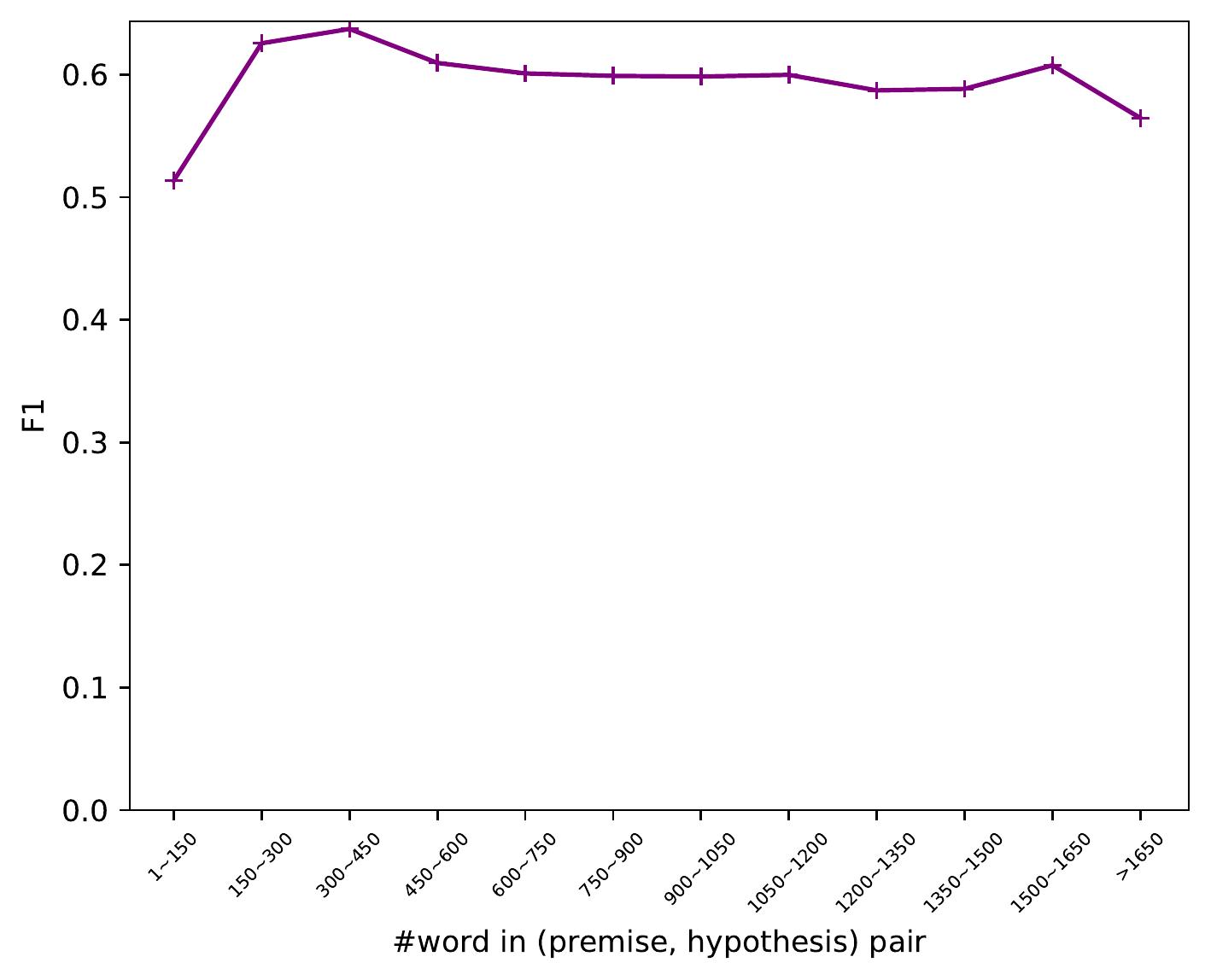}
}
\subfigure[F1 vs. hypothesis length] { \label{fig:Docf1hypo}
\includegraphics[width=7.5cm]{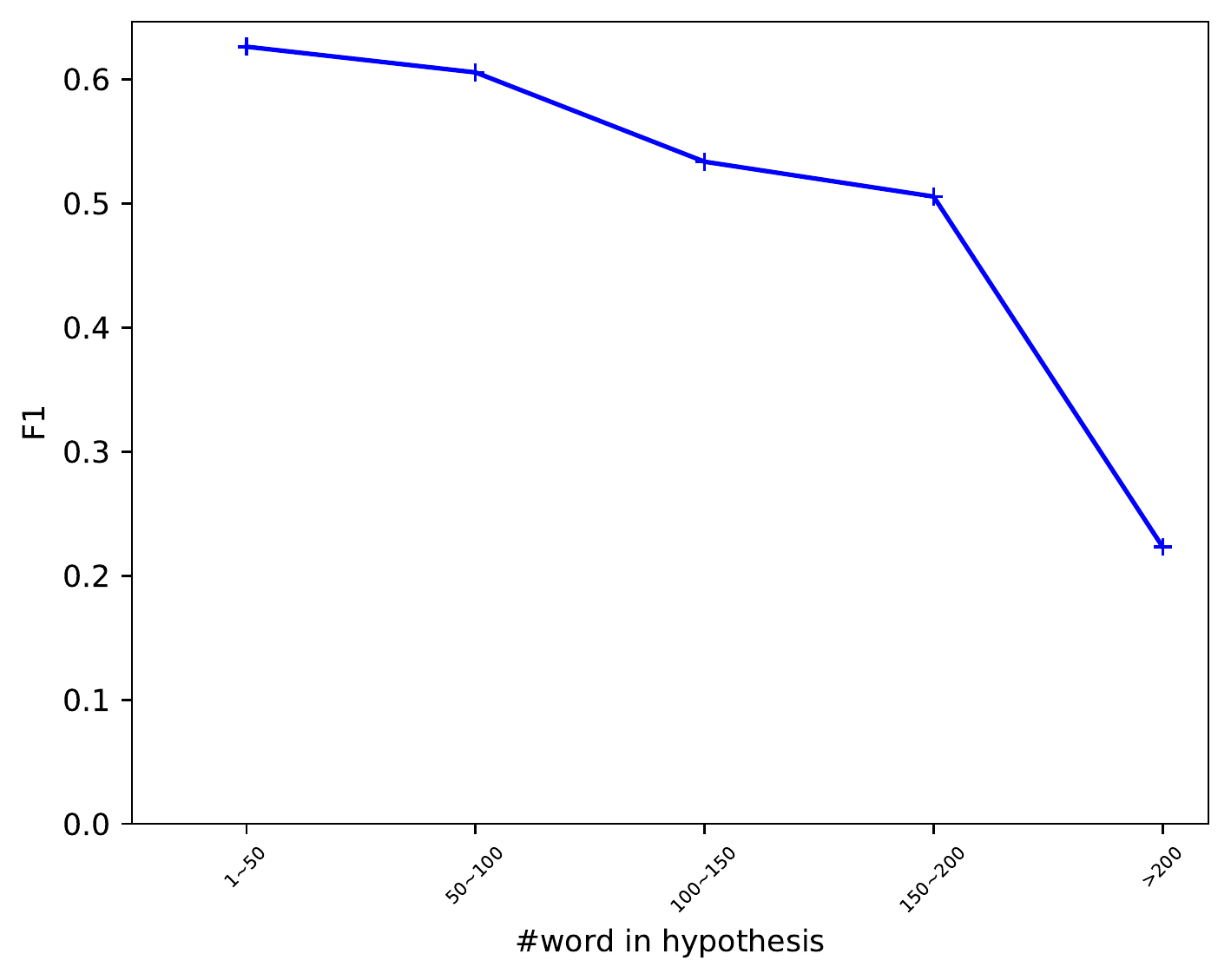}
}
\caption{Fine-grained F1 scores for different lengths of \dataname\enspace pairs or hypotheses alone.}
\label{fig:finegrainedf1}
\end{figure*}


\section{Experiments}
We  study three questions. (\textbf{Q}$_1$) How challenging is \dataname\enspace (especially with regard to different lengths of hypotheses)?   (\textbf{Q}$_2$) Out-of-domain evaluation, in which we train a system given \dataname\enspace and test it on  downstream NLP tasks that are not covered by the source tasks in \dataname\enspace construction. (\textbf{Q}$_3$) Could a system trained on \dataname\enspace work well on sentence-level NLI?

\subsection{The \dataname\enspace task is challenging}
The state-of-the-art systems on sentence-level NLI problems are largely based on transformers \cite{DBLPVaswaniSPUJGKP17},
such as BERT, RoBERTa \cite{DBLP11692}, etc. However, they can only handle maximal 512 tokens preprocessed by the WordPiece tokenizer \cite{DBLPWuSCLNMKCGMKSJL16}. This is an issue to build an effective document-level NLI machine. Therefore, for the main experiments, we also report Longformer \cite{DBLP05150} -- a RoBERTa variant that can handle up to 4096 tokens.  Longformer has two versions, one is ``Longformer-base'', the other is ``Longformer-large''. We currently  only report ``Longformer-base'' due to memory constraints. 

To answer the  question (\textbf{Q}$_1$), we compare the following systems  (we can include more baselines, but most popular approaches either are too weak or can only handle short piece of texts):

\textbullet\enspace \textbf{Hypothesis-only.} We train RoBERTa on hypotheses only.

\textbullet\enspace \textbf{RoBERTa-large.} Although we claimed that RoBERTa may not be a good platform to learn 
\dataname, here we report it  just for reference. Maximal token limit: 512 tokens. 

\textbullet\enspace \textbf{Longformer-base.} We use the released Longformer library\footnote{\url{https://github.com/allenai/longformer}} by \cite{DBLP05150}, training it on the full training set of \dataname, with length limit of 1.3K tokens, batch size 1 per GPU, and learning rate 5e-6.

All systems are trained for 5 epochs, and report the best model tuned on dev set. Table \ref{tab:results4task1} lists the F1 results of all systems on \dataname. We notice that ``hypothesis-only'' is just slightly higher than random guess, and is much lower than the ``RoBERTa-large'' system which takes both premises and hypotheses as input: 22.02 vs. 61.52 on test. Surprisingly, ``Longformer'''s performance is clearly below that of the RoBERTa, even if it covers more tokens, possibly because we do not have enough computing resources to fully explore the better settings of Longformer. Figure \ref{fig:token2f1} illustrates the impact of taking different numbers of tokens in Longformer, evaluated on dev set. In general, the more tokens the better performance. 

We further look at the fine-grained F1 reports on the various lengths of premise-hypothesis pairs and hypotheses alone. Figure \ref{fig:Docf1pair} shows that the system performance for pairs of lengths $>$ 450 does not change  clearly. This is probably  due to those models' truncation when the (premise, hypothesis) pairs are overlong (note that one word may be split into multiple tokens by the WordPiece tokenizer). 
Figure \ref{fig:Docf1hypo} demonstrates that the task gets increasingly challenging when the hypotheses become longer, which matches our intuition.

Overall, \dataname\enspace is a very challenging task that seeks solutions equipped with a stronger capability of representation learning.


\subsection{Applying \dataname\enspace to end NLP tasks}
To answer the question (\textbf{Q}$_2$), we play \dataname\enspace to see if it can help downstream NLP tasks. As \dataname\enspace is derived from summarization and QA already, we do not consider these two types of NLP tasks any more (since improvements on them are not surprising), especially when their domains are covered in \dataname. In addition, we have to explore tasks that have NLI-format data available --- converting an open NLP task to NLI format is not trivial and is beyond the scope of this work. Therefore, we consider the following two  NLP tasks:

\paragraph{FEVER \cite{DBLPhorneVCM18}.} FEVER is a benchmark dataset for fact-checking. Given an declarative sentence (aka. ``claim''), the task searches for textual evidences from Wikipedia articles and then decide the truth value of this sentence (i.e., support / refute / not-enough-info). 

We use the NLI-version of FEVER, released by \cite{DBLPNieCB19}: claims are hypotheses; premises corresponding to ``support'' or ``refute' claims  consist of  ground truth textual evidence and some other randomly sampled evidence; premises for  ``not-enough-info'' claims  are the concatenation of all selected evidential sentences by a previous SOTA fact-checking system. We combine ``refute'' and ``not-enough-info'' as a single class ``not\_entail'', and rename this data as ``FEVER-binary''. We randomly split FEVER-binary by  203,152/8,209/10,000 for train/dev/test respectively.\footnote{Please note that this data released by \cite{DBLPhorneVCM18} is different from the one used in FEVER leaderboard.}

\paragraph{MCTest \cite{DBLPichardsonBR13}.} 
In Related Work, we have introduced MCTest. Briefly, it is a multi-choice QA benchmark in the domain of \textbf{fictional story}.
The authors of MCTest released an NLI-version MCTest by combining the question and the positive (resp. negative) answer candidate as a positive (resp. negative) hypothesis. 

MCTest consists of two subsets. MCTest-160 contains 160
items (70 train, 30 dev, 60 test), each consisting of
a document, four questions followed by one correct
answer and three incorrect answers and MCTest-500
500 items (300 train, 50 dev, 150 test). \textit{MCTest has pretty limited labeled data; thus, it is a good testbed to investigate  \dataname\enspace in studying annotation-scarce tasks.} The MCTest has two official metrics: accuracy and NDCG (Normalized Discounted Cumulative Gain). Here we only report accuracy.

\begin{table}[t]
  \centering
  \begin{tabular}{ll|ccc}
& & FEVER  & \multicolumn{2}{c}{MCTest}\\
& &  binary & v160 & v500\\\hline\hline
& random & 50.00 & 25.00 & 25.00\\\hline
\multirow{3}{*}{\rotatebox{90}{pretrain}} &  MNLI & 86.64 & 75.41 &  70.66 \\
& ANLI &87.51 & 82.50 &78.66 \\
& \dataname & 88.84 & 90.00 & 85.83\\\cdashline{1-5}
& \enspace\enspace +finetune &\textbf{89.44} & \textbf{90.83} & \textbf{90.66}\\\hline

\multicolumn{2}{l|}{Prior state-of-the-art} & -- &  80.00 & 75.50 \\
\hline\hline
\end{tabular}
\caption{Train on \dataname, test on  NLP tasks that are out-of-domain  and require document-level NLI. SOTA of MCTest comes from \cite{DBLPYuZY19}.}\label{tab:testondownstreamNLP}
\end{table}

In this section, we still use RoBERTa-large and compare our \dataname\enspace with a latest NLI dataset ANLI in which the premises are longer than single sentences, and MNLI, the most widely-used sentence-level NLI dataset. For each data set (i.e., MNLI, ANLI or \dataname), we try two settings: (i) Using the data for pre-training, then do inference on FEVER-binary or MCTest directly without task-specific fine-tuning; (ii) First pre-training on the data, then fine-tune on FEVER-binary or MCTest. 

In Table \ref{tab:testondownstreamNLP}, \dataname\enspace can consistently  generalize better than ANLI and MNLI on the two NLP tasks FEVER-binary and MCTest. We notice that the pretrained model on \dataname\enspace demonstrates very strong performance on the two end tasks, even without any fine-tuning on the task-specific examples. Especially for MCTest, both the ``\dataname\enspace(pretrain)'' and ``\dataname+finetune'' surpass the prior state-of-the-art by large margins.



\subsection{Applying \dataname\enspace to sentence-level NLI}
To answer the question (\textbf{Q}$_3$), we use SciTail and MNLI as target sentence-level NLI tasks. SciTail is from the science domain with two classes ``entail'' and ``not\_entail''  split 23,596/1,304/2,126 (train/dev/test).  MNLI covers a   broad range of genres with three classes ``entail/neutral/contradict''  split 392,702/20k/20k (train/dev/test). Since the gold labels of the test set in MNLI  are not publicly available and \dataname\enspace is a binary classification task, we first unify  MNLI's ``neutral'' and ``contradict'' into ``not\_entail'', then build a new labeled test set by randomly sampling 13k from the original dev set (the remaining examples are the new dev set). So now we have train/dev/test of size 372,702/6,647/13k. We first try some popular Transformer-style models, such as BERT-large, RoBERTa-large and Longformer-base to check how much we can get by training a supervised system on the full training data. Afterwards, we build a classifier by training RoBERTa-large on \dataname\enspace with or without SciTail/MNLI-specific fine-tuning.  

\begin{table}[t]
\setlength{\tabcolsep}{3pt}
  \centering
  \begin{tabular}{l|cc}
& SciTail  & b-MNLI\\\hline\hline
majority & 60.33 & 66.66\\
ESIM \cite{DBLPChenZLWJI17} & 70.60 & -- \\
De-Att \cite{DBLPParikhT0U16} & 72.30 &  --  \\
DGEM \cite{DBLPKhotSC18} & 77.30 & -- \\
BERT-large & 89.71 & 90.55\\

Longformer-base & 92.23 & 92.03\\
RoBERTa-large &95.13 &93.95 \\\hline
\dataname\enspace (pretrain)& 78.17  & 91.13 \\
\enspace\enspace+finetune&  96.04 & 94.07 \\\hline
Prior state-of-the-art &97.70 & --\\
\hline\hline
\end{tabular}
\caption{Train on \dataname, test on sentence-level NLI benchmarks with or without fine-tuning. The SOTA of SciTail was reported by the DeBERTa model \cite{DBLP03654}.}\label{tab:testondownstreamNLI}
\end{table}

Table \ref{tab:testondownstreamNLI} shows that: (i) The pretrained model on \dataname\enspace indeed can generalize to some extend on both SciTail and MNLI. In particular, it gets SciTail accuracy 78.17  which is even higher than some task-specific fully-supervised models such as ``ESIM'', ``De-Att'' and ``DGEM''. The same pretrained system can also get comparable performance with BERT, Longformer and RoBERTa on binary-MNLI; this should be attributed to the strong generalization of ANLI towards MNLI \cite{DBLPabs191014599}; (ii) When do task-specific fine-tuning, our model can  further improve the performance and get very close to the state-of-the-art in SciTail.

\section{Summary}
In this work, we collect and release a large-scale document-level NLI dataset \dataname. It covers multiple genres and multiple ranges of lengths in both premises and hypotheses. We expect this dataset can help to solve some NLP problems that require document-level reasoning such as QA, summarization, fact-checking etc. In experiments, we show that \dataname\enspace can yield a model  generalizing well to downstream NLP tasks and some popular sentence-level NLI tasks. 

\section*{Acknowledgments}

The authors would like to thank   Nitish Shirish Keskar, our colleague at Salesforce Research, for helping play the CTRL code, and thank the anonymous reviewers for insightful comments and suggestions. 

\bibliographystyle{acl_natbib}
\bibliography{anthology,acl2021}


\end{document}